# *UnMA*-CapSumT: Unified and Multi-Head Attention-driven Caption Summarization Transformer

[1]Dhruv Sharma, [2]Chhavi Dhiman, [3]Dinesh Kumar

[1,2,3]Deapartment of Electronics and Communication Engineering,
Delhi Technological University, Delhi, India

[1]*dhruv.0906@yahoo.in*, [2]*chhavi.dhiman@dtu.ac.in*, [3]*dineshkumar@dtu.ac.in*

*Abstract—* Image captioning is the generation of natural language descriptions of images which have increased immense popularity in the recent past. With this different deep-learning techniques are devised for the development of factual and stylized image captioning models. Previous models focused more on the generation of factual and stylized captions separately providing more than one caption for a single image. The descriptions generated from these suffer from out-of-vocabulary and repetition issues. To the best of our knowledge, no such work exists that provided a description that integrates different captioning methods to describe the contents of an image with factual and stylized (romantic and humorous) elements. To overcome these limitations, this paper presents a novel Unified Attention and Multi-Head Attention-driven Caption Summarization Transformer (*UnMA*-CapSumT) based Captioning Framework. It utilizes both factual captions and stylized captions generated by the Modified Adaptive Attention-based factual image captioning model (MAA-FIC) and Style Factored Bi-LSTM with attention (SF-Bi-ALSTM) driven stylized image captioning model respectively. SF-Bi-ALSTM-based stylized IC model generates two prominent styles of expression- {romance, and humor}. The proposed summarizer $UnMHA-ST$ combines both factual and stylized descriptions of an input image to generate styled rich coherent summarized captions. The proposed $UnMHA-ST$ transformer learns and summarizes different linguistic styles efficiently by incorporating proposed word embedding fastText with Attention Word Embedding (fTA-WE) and pointer-generator network with coverage mechanism concept to solve the out-of-vocabulary issues and repetition problem. Extensive experiments are conducted on Flickr8K and a subset of FlickrStyle10K with supporting ablation studies to prove the efficiency and efficacy of the proposed framework. The code of the proposed work is available at: https://github.com/dhruvsharma09/UnMA-CapSumT.git

*Index Terms—* computer vision, factual image captioning, natural language processing, stylized image captioning, text summarization, word embedding,

# 1. INTRODUCTION

Image captioning [1] [2] identifies objects within an image, comprehending their interactions to describe the contents of an image in the form of descriptive sentences. Caption Generation is a very challenging task as it generates descriptive and procedural texts that are more effective for image comprehension. Hence, it is employed for numerous applications like human-robot interaction, aid to the blind [3] [4], visual question answering [5], etc. Most studies [6] [7] [8] on image captioning primarily focus on the description of the factual content of an image while ignoring the underlying emotion or style of the same image. On the other hand, very few studies [9] [10] learned the knowledge of factual content and its linguistic style for the generation of style-based descriptions of images. This on the other hand facilitates many real-world applications, like chatbots, or enlightening users in photo captioning for different social media platforms, storytelling, etc. as it greatly enriches the expressibility of the caption, making it more attractive. Further, style-based caption generation not only describes the contents of an image but also analyzes the intrinsic style-based elements in the captioning [11] [12]. Also, this requires paired image-sentiment [11] caption data and word-level supervision to emphasize the sentiment words, therefore making style-based captioning very expensive.

With the advancements in deep-learning technology, different models for factual [7] [8] and stylized image captioning [11] [12] are developed. These works generate separate sentences for factual and stylized (romantic and humorous) descriptions of an image. Therefore, these methods sometimes may give erroneous descriptions by either describing the factual content in dull language or may lead to the incorrect representation of the style. This may be due to poor learning knowledge of factual content and its associated linguistic styles. To the best of our knowledge, there are no such works in literature that provide single-sentence summarized descriptions incorporating factual, romantic, and humorous content.

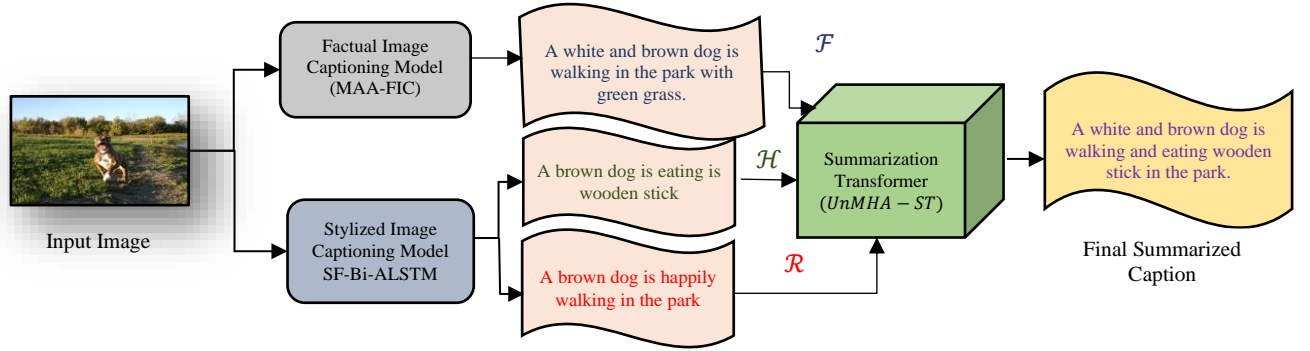

**Fig. 1:** Block Diagram Representation of the proposed $UnMA-$CapSum Transformer based Captioning Framework

To address the abovementioned issues and to improve the learning of knowledge of factual content and its associated linguistic styles this paper extended the task of image captioning to abstractive text summarization. The task is to automatically summarize the source article into an accurate short version that reflects its central content comprehensively. Though the summaries generated also suffers from major issues of rare words that arises due to repetition of words or due to out-of-vocabulary issues [13] [14]. In this direction, this paper presents a novel Unified Attention (Un-A) and Multi-Head Attention-driven Caption Summarization Transformer ($UnMA$-CapSumT) based Captioning Framework. An overview of the proposed framework is depicted in Fig. 1, the framework is twofold: $(i)$ it integrates MAA-FIC-based factual image captioning and SF-Bi-ALSTM based stylized image captioning techniques for the generation of factual, stylized – {romantic, and humorous} description of an image, and $(ii)$ the proposed summarizer $UnMHA-ST$ combines both factual and stylized descriptions of input image to generate styled rich coherent summarized captions.

The main contributions of this work are:

i. This paper proposed a Unified and Multi-head Attention-based Caption Summarization Transformer $UnMHA-ST$ to capture the intra- and intermodal interactions of multimodal information and generates refined attended representations. Also, it utilizes the concept of a pointer generator network and coverage mechanism to solve the problems of rare words which arise due to out-of-vocabulary (OOV) and repetition issues.
ii. The proposed framework integrates Modified Adaptive Attention-based (MAA-FIC) and SF-Bi-ALSTM-based factual and stylized image captioning models for the generation of factual and stylized descriptions of images which are utilized by $UnMHA-ST$ for generation of styled rich coherent summarized captions.
iii. The paper also presents an efficient Attention enabled fastText word embedding, fTA-WE, that integrates the attention mechanism into the Continuous Bag of words (CBOW) model of fastText that efficiently learns vector representation of words and enhances the performance of the proposed $UnMHA-ST$.
iv. Extensive experiments on Flickr8K [15], and a subset of FlickrStyle10K [9] demonstrate the efficiency and efficacy of the proposed $UnMA$-CapSumT framework.

The organization of the rest of the paper is as follows: Section 2 presents the different state-of-the-art for factual image captioning, stylized image captioning, and text summarization. Section 3 discusses the proposed methodology. Section 4 presents the Dataset Used, Implementation details with the results obtained. Finally, Conclusions are drawn in Section 5.

## 2. RELATED WORKS

This section presents different techniques from past for factual image captioning, stylized image captioning, and text summarization.

## 2.1 Factual Image Captioning

Traditional image captioning methods [16] [17] [18] [19] fail for many practical applications as they are unable to generate variable-length descriptions and the descriptions generated are less natural. Analysis of complex, high-dimensional, and noise-contaminated data is extremely difficult. Therefore, it is essential to create new algorithms that can categorize, summarize, extract, and transform data into understandable forms. Hence, deep-learning-based models have demonstrated exceptional achievements in the past few years. To overcome these limitations deep learning methods for image captioning were encouraged. These methods [1] [20] rely on large-scale datasets which efficiently and effectively generate variable-length descriptions of images.

Deep-learning-based models have demonstrated exceptional achievements in the past few years. These methods [20] [21] proposed a generative model that combined recent advances in CV and machine translation which is trained to maximize the likelihood of the target description sentence for a given training image. In addition to this, [20] directly models the probability distribution of generating a word given previous words and an image. To generate more natural and descriptive long-length sentences, guided-LSTM or g-LSTM [22] is proposed. g-LSTM adds semantic information extracted from the image as extra input to each unit of the LSTM block. For the generation of captions for static as well as dynamic images, Donhaue et al. [23] stacked multiple LSTMs to jointly train and learn temporal dynamics and convolutional perceptual representations. Wang et al. [24]generated semantically rich descriptions of images using deeper Bi-directional LSTM framework. This framework utilized past and future context information to learn long-term visual language interactions.

The above-discussed methods provide captions for the entire scene rather than taking into account the spatial elements of an image that are important for captioning. To overcome these limitations soft-attention [1], spatial and channel-wise attention [6], adaptive [7] [25] [26], multi-head and self-attention [27], and Semantic attention [28] based captioning of images came into existence. The attention-based methods [20] [21] have successfully surpassed the performance of traditional deep-learning models by highlighting salient parts of the image and by providing higher-order interactions between objects and the scene. However, to concentrate more on the visual salient regions Xiao et al. [29] proposed Dual-LSTM and adaptive attention mechanism-based model for image captioning and visual question answering. The proposed model flexibly makes a trade-off between the visual semantic region and textual content to achieve a significant performance of prior state-of- the-art approaches. Yan et al. [30] proposed a task-adaptive attention module that introduced diversity regularization to enhance the expression ability of the attention mechanism. Further, Xiao et al. [31] presented a new variant of LSTM, named ALSTM (Attention-LSTM) for image captioning for generation of captions by paying more attention to the most relevant context words.

## 2.2 Stylized Image Captioning

Stylized or style-based image captioning is the description of images with different styles or sentiments. Stylized image captioning when compared with factual image captioning provides a better understanding of images. Ideally, a stylized image captioning model should fulfill two conditions, (i) provides the correct description of images, and (ii) generates correct stylized words or phrases in the correct positions of descriptions. To fulfill these conditions, a stylized image captioning model utilizes non-parallel [9] [12] [32] [33] or parallel [10] [11] [34] [35] stylized image captioning data. Mathews et al. [34] described an image using positive and negative sentiments. The model proposed utilized [34] word-level regularization for producing emotional image captions. Another work [32] developed a novel semantic term representation to disentangle content and style in descriptions. Gan et al. [9] proposed a novel factored LSTM which automatically distills the style factors in the monolingual text corpus and generated romantic and humorous captions for images. To efficiently describe the visual contents of an image using linguistic styles very few works [10] [35] focused on splitting the stylized sentence into ($i$) style-related part that reflects the linguistic style and ($ii$) a content-related part that contains the visual content. To generate controllable stylized descriptions of images Chen et al. [11] suggested adaptive learning with attention. Also, it provided more stylish image descriptions relevant to the content of the images. MSCap [12] is the first attempt toward the generation of image descriptions in multiple styles with a single model. Zhao et al. [33] proposed a novel model with a memory module that memorizes the knowledge of linguistic style and distills the content-relevant style knowledge with an attention mechanism for generating stylized descriptions. Further, Li et al. [36] proposed a sentimental image captioning to reflect the inherited sentiments of an image. The methods discussed above focus on the generation of image descriptions with linguistic styles in a single decoding process. In contrast, the proposed *UnMA* CapSumT generates a single summarized image description with factual and stylized elements.

## 2.3 Text Summarization

Researchers in the past have developed many traditional techniques or methods for summarization of text, such as feature score [37], semantic information [38], template [39], classification [40], and dynamic programming [41]. These methods generated incomplete, redundant summaries for the input text data. Also, these methods are easily affected by abnormal data which leads to the generation of uncorrelated text at the output. To overcome the generation of incomplete, redundant, and uncorrelated text at the output deep-learning-based automatic abstractive summarization methods are encouraged. Rush et al. [42] proposed a fully data-driven attention-based approach by combining probabilistic models and generation algorithms to produce accurate summaries. Nallapati et al. [13] [14] described summarization tasks using attentional encoder-decoder RNN and provided state-of-the-art results. The abstractive summaries generated by [42] [13] [14] suffer from OOV and low-frequency issues. To overcome these limitations text summarization with a pointer-generator network and coverage mechanism [43] was proposed. This solves the issues related to OOV and also avoids the generation of repetitive and incoherent phrases in summarized text. Hsu et al. [44] introduced a novel inconsistency loss function for abstractive and extractive text summarization with a pointer-generator network and coverage mechanism. Inconsistency loss penalized the inconsistency between two levels of attention and efficiently solves the issues related to OOV and repetitive phrases in the generated summary.

Li et al. [45] proposed a multi-head attention-driven abstractive text summarization network that learned relevant information in different representation subspaces. Yao et al. [46] proposed dual-encoder-based abstractive text summarization in which the primary encoder conducts coarse encoding and the second one models the importance of words. In recent past transformer-based architectures are now being encouraged for summarization tasks. In this direction, Deaton et al. [47] proposed a transformer-based approach for abstractive summarization tasks. The model [47] incorporated basic transformer-based architecture [48] with a pointer-generator network and coverage mechanism. Further, this paper presented multi-head attention and unified attention-driven $UnMHA - ST$ transformer model with a pointer-generator network and coverage mechanism. this provides improvements in the performance of the summarization task by reducing redundancy and factual errors.

## 3. PROPOSED WORK

This section discusses the proposed model, which is divided into two phases i.e., image captioning, and text summarization. In the proposed model, as depicted in Fig. 2, two image captioning techniques are integrated to generate factual, romantic, and humorous captions of an image. Further, these captions are fed to the transformer-based text summarization model to provide the final summarized caption which contains all three (factual, romantic, and humorous) elements in a single caption.

### 3.1 Modified Adaptive Attention-based Factual Image Captioning Model (MAA-FIC)

This section describes in detail the proposed MAA-FIC factual image captioning model. The input to the model is an image $\mathcal{I}$ and the output is the descriptive factual sentence $\mathcal{F}$ with $\hbar$ encoded words: $\mathcal{F} = \{w_1, w_2, \ldots .. w_\hbar\}$.

#### 3.1.1 Feature Extraction

For object detection, the proposed MAA-FIC model incorporates Faster R-CNN for the detection of objects. After the detection of objects, each detected object is mapped to a feature vector. For each image $\mathcal{I}$, $n$-objects are detected, given by, $\{\Phi_1, \Phi_2, \ldots, \Phi_n\}$; $\Phi_i \in \mathbb{D}^d$ where, $\mathbb{D}^d$ is the $\mathbb{D} -$

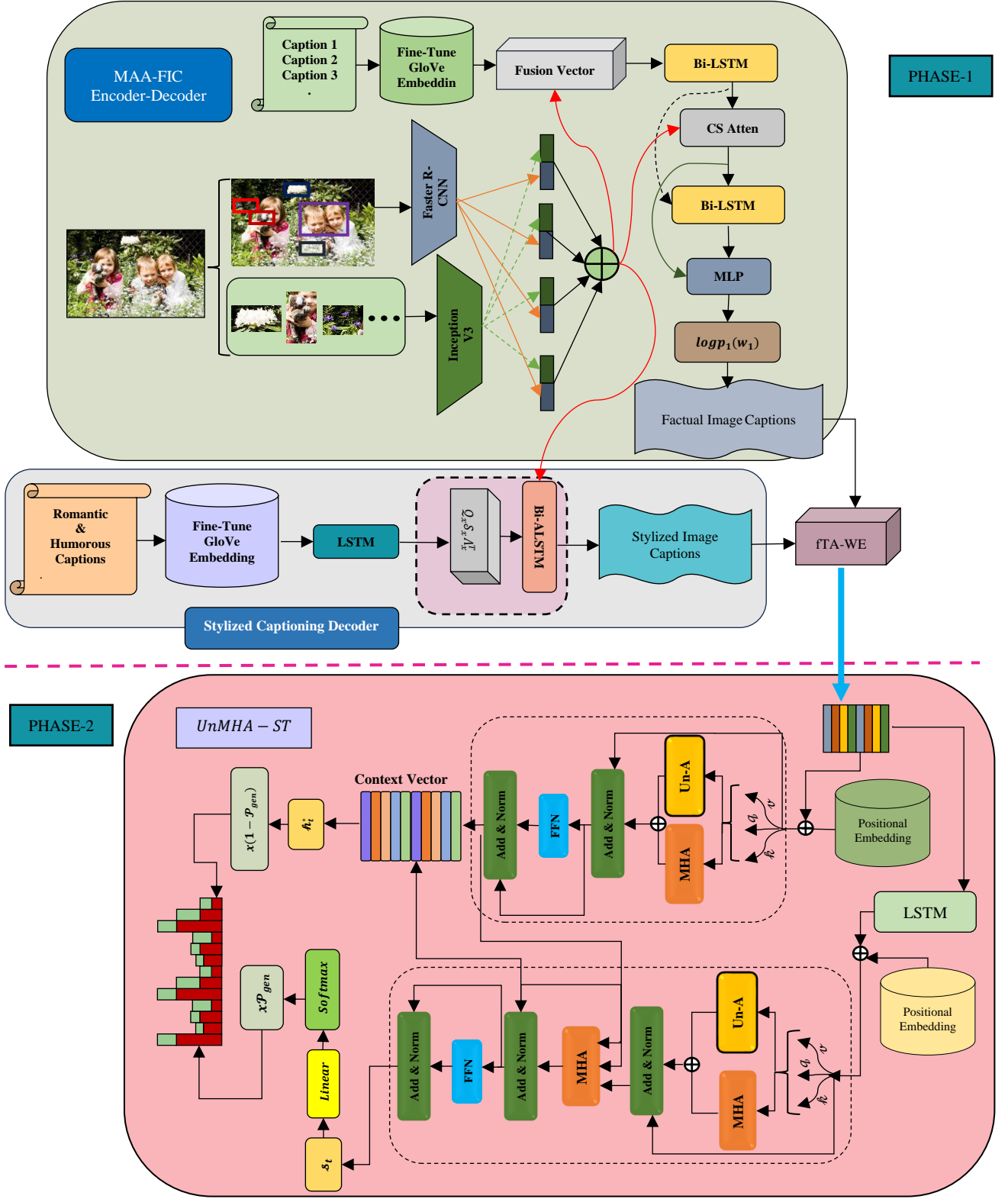

Fig. 2: Proposed Methodology: Phase-1: MAA-FIC module and Stylized Image Captioning Module, Phase 2: $UnMHA - ST$: Unified and Multi-head Attention based Summarization Transformer

dimensional vector of top $n$-boxes as the region of objects.

The localization of objects is performed to extract spatial relationships between objects. The $n$-objects detected using Faster R-CNN [49] for each image are fed to Inception-V3 [50] and output the feature vector that represents the spatial location of each object, represented as $\{\theta_1, \theta_2, ..., \theta_n\}$; $\theta_i \in \mathbb{D}^{\mathbb{T}}$, where, $\mathbb{D}^{\mathbb{T}}$ is the $t - dimensional$ vector of the spatial location of each object. The features extracted from the Faster R-CNN module and Inception-V3 module are concatenated. Therefore, each concatenated feature vector consists of a detected object feature vector ($\Phi_i$) and the feature vector obtained from the localization of objects ($\theta_i$). Mathematically, the concatenated image feature vector is represented as:

$$\mathbb{F}_i = [\Phi_i; \theta_i], \mathbb{F}_i \in d^D, D = d + \mathbb{T} \qquad (1)$$

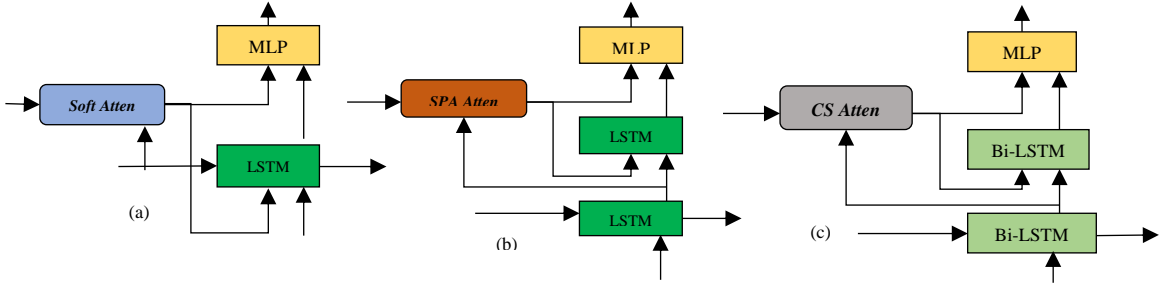

**Fig. 3:** (a) Soft-attention Mechanism, (b) Adaptive Attention Mechanism, and (c) Proposed Modified Adaptive Attention

*3.1.2 Text Generation Network*

This section introduces Modified Adaptive based Attention ($MAA$) and Bi-LSTM-based factual language decoder module. The $MAA$ utilizes both channel and spatial attention mechanism with bidirectional sequence learning. From the definition of attention mechanism, it utilizes the most relevant part of the input in a flexible manner. This technique combines the weighted combination of all the encoded input vectors, with the most relevant vectors being attributed with the highest weights. For the model with attention mechanism:

$$a_t = f(\mathbb{F}, \hbar_t) \qquad (2)$$

where $f$ is the attention mechanism and $\mathbb{F} = [\mathbb{F}_1, \mathbb{F}_2 ..., \mathbb{F}_k]$ represents the concatenated image features and $\hbar_t$ is the hidden state of the RNN at time $t$. Channel-wise attention distribution for $k$ regions is defined as:

$$\tilde{\mathcal{C}} = \tilde{\mathcal{C}}_{\mathcal{W}} = \mathcal{W}_{hc}^T \tanh((\mathcal{W}_{c\mathbb{F}} \otimes \mathbb{F} + \beta_c) \oplus (\mathcal{W}_{fc}\hbar_t)\mathcal{I}^T) \qquad (3)$$
$$\psi = softmax(\tilde{\mathcal{C}}) \qquad (4)$$

Further, spatial attention weights of the $k$ regions mathematically expressed as:

$$\tilde{\mathcal{S}} = \tilde{\mathcal{S}}_{\mathcal{W}} = \mathcal{W}_{hs}^T \tanh((\mathcal{W}_{s\mathbb{F}}\mathbb{F} + \beta_s) \oplus (\mathcal{W}_{fs}\hbar_t)\mathcal{I}^T) \qquad (5)$$
$$\varphi = softmax(\tilde{\mathcal{S}}) \qquad (6)$$

where, $\beta_s$ and $\beta_c$ are bias terms. After attaining the channel and spatial attention weights, channel-spatial weights are thus combined to obtain a modulated feature map. This provides the modified attentional distributions for the $k$ regions as:

$$\psi = \phi_c(h_t, \mathbb{F}) \qquad (7)$$
$$\varphi = \phi_s(h_t, f(\mathbb{F}, \psi)) \qquad (8)$$
$$\tilde{\mathcal{X}} = f(\mathbb{F}, \psi, \varphi) \qquad (9)$$

Therefore, the final modified attention weights are defined mathematically as:

$$a_t = \sum_{i=1}^{k} (\mathcal{X}_{it}\mathbb{F}_{it}) \qquad (10)$$

Based on the channel and spatial attention mechanism, modified adaptive attention is proposed with visual sentinel and Bi-LSTM layers. Fig. 3 presents the basic architecture of MAA. The visual sentinel helps the model to focus more on wither image information or on the language rules. The memory unit in Bi-LSTM stores the information of both the previous partial visual information and the language rules. Therefore, visual sentinel $s_t$ formula based on the Bi-LSTM memory unit is:

$$m_t = \theta(\mathcal{W}_x \mathcal{X}_t + \mathcal{W}_h h_{t-1}) \qquad (11)$$
$$s_t = m_t \odot \tanh(\tilde{h}_t) \qquad (12)$$

Where, $\mathcal{X}_t$ is the input of the Bi-LSTM, $m_t$ is the sentinel gate, is the output of the last time node $h_{t-1}$), and $\tilde{h}_t$ is the memory cell. Also, the modified adaptive attention (MAA) mechanism is obtained following the sentinel gate $s_t$

$$\hat{a}_t = \delta_t s_t + (1 - \delta_t)a_t \qquad (13)$$

$\delta_t \in [0,1]$ and is the new sentinel gate at time t. The value of $\delta_t$ decides whether the model focuses on image information or language rules. If the value of $\delta_t$ is equal to zero, it denotes focus on image information whereas the value of 1 indicates that the focus is on the language rules. The channel-spatial attention distribution of $k$ regions obtained by splicing an element and is given by:

$$\widehat{\mathcal{X}} = softmax([\widetilde{\mathcal{X}}; \mathcal{W}_h^T \tanh(\mathcal{W}_s s_t + \mathcal{W}_m h_t))]) \qquad (14)$$

To generate the natural language description of an input image, fine-tuned GloVe text embeddings are employed and are represented as $\mathbb{T}$. The input to the text generation module is the weighted fused vector obtained after the fusion of image features $\mathbb{F}$ and text features $\mathbb{T}$.

$$\mathcal{Q}_t = \{\mathbb{F}_t \mathbb{T}_t\} \qquad (15)$$

The fused feature vector $\mathcal{Q}_t$ is incorporated for the generation of accurate description of an image. Further, the resulting output generates a word on the next time node using the MLP function that incorporates backpropagation through a time algorithm to update the parameters of the LSTM network. The objective function defined for the optimization of the proposed model is given by:

$$\Delta = arg \max_{\vartheta} \sum_{r,y} \sum_{t=0}^{N} \log p(\mathcal{F}_t | r, \Delta, \mathcal{F}_1, \mathcal{F}_2 \ldots \mathcal{F}_{t-1}) \qquad (16)$$

$\vartheta$ is the learnable parameter with $m$ feature maps whose weight is $r \in \{r_1, r_2 \ldots r_t\}$. The cross-entropy loss is incorporated to minimize the loss function and to maximize the probability of each correct word appearing. The cross entropy-loss function is given by:

$$\mathcal{L}(\Delta) = -\sum_{t=1}^{N} \log(p_t(w_t^* | \mathbb{G}_{1:t-1}^*)) \qquad (17)$$

here $N$ represents total words in a sentence; $\Delta$ denotes all the parameters in the model; $\mathbb{G}_{1:t-1}^*$ defines the ground truth.

## 3.2 SF-Bi-ALSTM Based Stylized Image Captioning

This section introduces the SF-Bi-ALSTM module, which serves as the building block for the generation of style-based descriptions of images. The input to the proposed style-based caption generation model is image $\mathcal{J}$ and the output is the descriptive romantic and humorous sentences $\mathcal{R}$ and $\mathcal{H}$ with $n$ and $m$ encoded words: $\mathcal{R} = \{w_1, w_2, \ldots w_n\}$, $\mathcal{H} = \{w_1, w_2, \ldots w_m\}$. The proposed stylized captioning model is based on encoder-decoder-based architecture. For the encoder, this model employs the same strategy as employed by the MAA-FIC model as discussed in *Section A* whereas the SF-Bi-ALSTM model is utilized for the generation of style-based captions.

### 3.2.1 SF-Bi-ALSTM Module

This section presents a new and modified variant of LSTM and Factored LSTM namely, Modified Style Factored Bi-LSTM. Traditional LSTM captures the long-term dependencies among words in sentences but fails to capture the styles in sentences. Also, Factored LSTM incorporated style factors into the visual caption generation model but fails to generate more attractive and coherent style-based descriptions. To overcome these limitations SF-Bi-ALSTM Module is designed that produces more meaningful stylized descriptions, combining LSTM with attention layers from both directions. Further, the proposed Bi-LSTM learns to refine the input vector from network hidden states and sequential context information.

Consider, $\mathcal{M}_x \in \mathbb{R}^{m \times n}$, then $\mathcal{Q}_x \in \mathbb{R}^{m \times r}$, $\mathcal{S}_x \in \mathbb{R}^{r \times r}$, and $\Lambda_x \in \mathbb{R}^{r \times n}$. $\mathcal{M}_x$ can be represented in the form of three matrices:

$$\mathcal{M}_x = \mathcal{Q}_x \mathcal{S}_x \Lambda_x^T \qquad (18)$$

Further, the style-specific matrix $\{\mathcal{S}_x\}$ can be represented as:

$$\mathcal{S}_x = \frac{1}{N} |\mathcal{S}\mathcal{S}^T| \qquad (19)$$

The memory gates and cells in the traditional Bi-LSTM are modified with respect to $\mathcal{M}_x$ and attention mechanism:

$$\tilde{x}_t = sigmoid(\mathcal{Q}_{x_t} \mathcal{S}_{x_t} \Lambda_{x_t} x_t + W_h \tilde{h}_{t-1}) \odot x_t \qquad (20)$$
$$\tilde{\iota}_t = sigmoid(\mathcal{Q}_{ix} \mathcal{S}_{ix} \Lambda_{ix} \tilde{x}_t + W_{ih} \tilde{h}_{t-1} + \tilde{b}_i) \qquad (21)$$
$$\tilde{f}_t = sigmoid(\mathcal{Q}_{fx} \mathcal{S}_{fx} \Lambda_{fx} \tilde{x}_t + W_{fh} \tilde{h}_{t-1} + \tilde{b}_f) \qquad (22)$$
$$\tilde{o}_t = sigmoid(\mathcal{Q}_{ox} \mathcal{S}_{ox} \Lambda_{ox} \tilde{x}_t + W_{oh} \tilde{h}_{t-1} + \tilde{b}_o) \qquad (23)$$

$$\tilde{g}_t = sigmoid(Q_{gx}S_{gx}\Lambda_{gx}\tilde{x}_t + W_{gh}\tilde{h}_{t-1} + \tilde{b}_g) \quad (24)$$
$$\tilde{c}_t = \tilde{f}_t \odot c_{t-1} + \tilde{x}_t \odot \tilde{g}_t \quad (25)$$
$$h_t = \tilde{o}_t \odot tanh\tilde{c}_t \quad (26)$$
$$p_{t+1} = (\mathbb{C}h_t) \quad (27)$$

where, $x_t$ is the input and an input update gate is also defined as $\tilde{x}_t$ according to input and the hidden state $h_{t-1}$. Also, $\Lambda_{ix}$, $\Lambda_{fx}, \Lambda_{ox}$, and $\Lambda_{gx}$ are the input weight matrices, and $W_{ix}$, $W_{fx}, W_{ox}$, and $W_{gx}$ are the weight matrices which are applied to recurrently update the matrices of hidden states. Also $\{\mathcal{M}\}, \{Q\}$, and $\{\Lambda\}$ are the matrices shared by two styles of romantic $S_\mathcal{R}$ and humorous $S_\mathcal{H}$ respectively. The style factored Bi-ALSTM model is implemented as:

$$h_t^f = \tilde{o}_t^f \odot tanh\tilde{c}_t^f \quad (28)$$
$$h_t^b = \tilde{o}_t^b \odot tanh\tilde{c}_t^b \quad (29)$$
$$y_t = W_{hy}^f h_t^f + W_{hy}^b h_t^b + b_y \quad (30)$$

*3.2.2 Style-Based Language Generation*

To generate the style-based descriptions of images the decoder of the proposed model leverages multi-task learning for sequence tagging [51]. This helps to learn to disentangle the style factors from the text corpus. The input to the SFA-Bi-LSTM module is the image feature matrix $\mathcal{I}_f$ extracted from the image encoder and the text feature embedding matrix $\mathcal{T}_f$ for romantic and humorous text corpus. The matrix $\mathcal{T}_f$ is obtained via fine-tuned GloVe embedding. The extracted features are utilized by the SFA-Bi-ALSTM module according to equations 20-30. The output generated from the proposed Bi-LSTM is soft-attended and generates either romantic or humorous descriptions of images. Also, at time step $t$, negative log-likelihood loss is incorporated to minimize the loss function and to maximize the probability of each word appearing for both romantic and humorous styles.

$$\mathcal{L}(\Theta) = -\sum_{i=1}^{n} s_{ij}log\hat{s}_{\Theta,ij} + (1 - s_{ij})\log(1 - \hat{s}_{\Theta,ij}) \quad (31)$$

C. Text Summarization Module

The image descriptions generated from the MAA-FIC model and SF-Bi-ALSTM based stylized captioning model are collected and summarized to generate a caption that contains factual, romantic, and humorous content in one caption. This section discusses about the proposed novel transformer-based summarization model, UnMHA-ST with *fTA-Word-Embeddings*, to generate a summarized caption for an image.

**3.3 fTA-Word-Embedding (fTA-WE)**

To improve the performance and interpretability of the word embedding models, this work proposes a fTA (fastText with Attention) word embedding as depicted in Fig. 4. The proposed embedding module utilizes fastText [52] word embeddings with soft-attention [53] to learn vector representation of words such that the words which are semantically related and close to each other. Also, fTA-WE leverage sub-words to enrich the vector space for rare and unseen words. The context vector $c_v$ for vocabulary $\mathcal{V} = [v_1, \ldots, v_\mathcal{N}]^T \in \mathbb{R}^{\mathcal{N} \times \mathcal{D}}$ of size $\mathcal{N}$ and word vector size $\mathcal{D}$ for fastText is represented by:

$$c_v = \sum_{i \in [-b,b]-\{0\}} u_{e_j} \quad (32)$$

where, $\mathcal{U} = [u_1, \ldots, u_\mathcal{N}] \in \mathbb{R}^{\mathcal{N} \times \mathcal{D}}$ represents the sub-words and is used in the calculations of context vectors to efficiently learn a representation for each word. Taking the word $'white'$ as an example the sub words formed for $n = 3\ grams$ are $< wh, whi, hit, \ldots >$. Further, $e_j$ and $b$ are the index of each word and the size of the context window. If $w_0$ is the masked word index with vector $v_{w_0}$, the probability of $w_0$ to occur in the context of $\{w_{-b}, \ldots, w_{-1}, w_1, \ldots, w_b\}$

$$p(w_0|W_{[-b,b]-\{0\}}) \propto expv_{w_0}^T c_v \quad (33)$$

The embedding of each word is represented by the embeddings of all sub-words which can be expressed mathematically as:

$$\tilde{u}_w = \sum_{e_j \in \mathbb{S}_w} u_{e_j} \quad (34)$$

The attention mechanism of the fTA-WE embeddings, utilizes key matrix $\mathcal{K}$, value $V$, and query matrix $Q$ the attention weight and the context vector with attention, defined as:

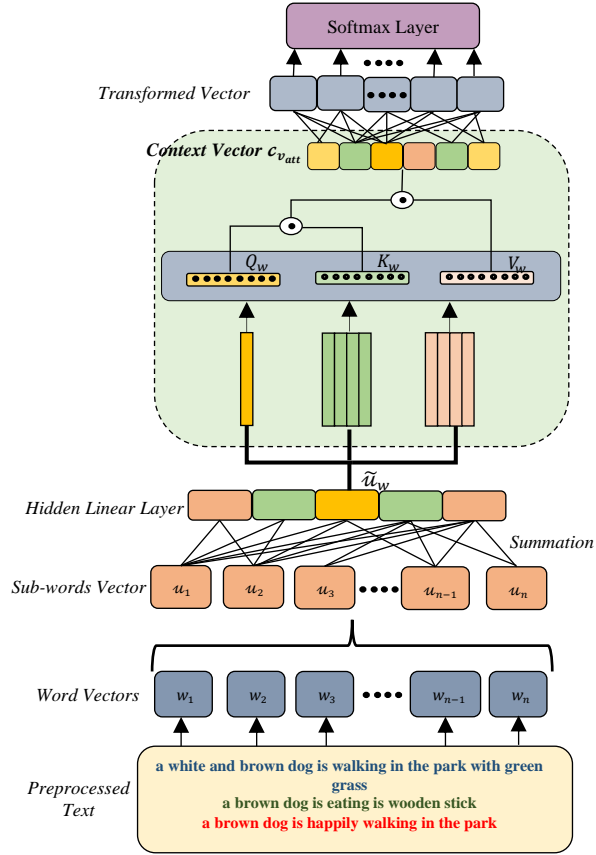

**Fig. 4**: Proposed fTA-WE

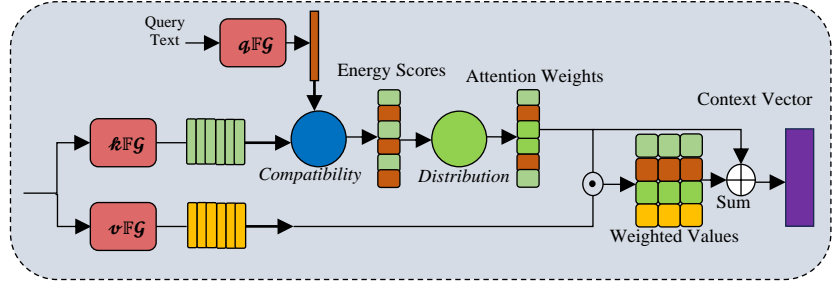

Fig. 5: Proposed Unified Attention Module

$$a_{w_i} = \exp\left(v_{w_0}^T c_v\right) \tag{35}$$

$$c_{v_{att}} = \sum_{i \in [-b,b]-\{0\}} a_{w_i} \left(\sum_{e_j \in \mathbb{S}_w} \tilde{u}_{e_j}\right) \tag{36}$$

The probability of the masked word is given by:

$$p(w_0 | W_{[-b,b]-\{0\}}) \propto \exp\left(\tilde{u}_{w_0}^T c_{v_{att}}\right) \tag{37}$$

The attended vector is linearly transformed to obtain the transformed vector representation of the input text. Furthermore, the obtained transformed vector representation is used as an input to the text summarization transformer to generate the summarized caption for the images.

### 3.4 $UnMHA - ST$ Text Summarization Transformer

This section provides a detailed description of the abstractive transformer-based text summarization model. The summarization model takes input from two captioning mechanisms discussed in section 3.1 and 3.2 respectively. It generates a summarized caption of an image that reflects the factual, romantic, and humorous contents in a single caption. The proposed $UnMHA - ST$ as depicted in Fig. 2, utilizes a unified attention mechanism in addition to multi-head attention. This helps in concurrently capturing the intra-modal and intermodal interactions of multimodal information which further helps in the generation of their related attended representations. With a unified attention module, the proposed $UnMHA - ST$ also leverages a pointer-generator network and coverage mechanism to solve the problems of rare words which arise due to out-of-vocabulary (OOV) and repetition issues. The input to the transformer network is the text embedding that is generated from the fTA-WE model. These embeddings are added with the positional embedding and are further split into key $k$, values $v$, and query $q$, respectively. To jointly attend information from different representation subspaces at different positions, MHA is employed which is mathematically represented as:

$$Attention(k, q, v) = softmax(\frac{qk^T}{\sqrt{d_k}})v \qquad (38)$$

$$MHA(k, q, v) = Concat(h_1, h_2, \dots, h_i)W^O \qquad (39)$$

$$\text{where, } h_j = Attention(kW_j^k, qW_j^q, vW_j^v) \qquad (40)$$

The unified attention (Un-A) module is depicted in Fig. 5 utilizes an iterative alternating attention mechanism [54] that allows a fine-grained exploration of both the query and the document. Un-A do not collapse the query into a single vector instead it evaluated keys and query by using a compatibility function $e = \mathbb{F}(q, k)$. The function $\mathbb{F}$ is the energy function [51]. The energy scores thus obtained are transferred into the attention weights using a distribution function $\mathcal{G}$, $\mathcal{E} = \mathcal{G}(e)$. Therefore, the vector obtained from the compatibility and distribution function is combined with the transformer values v, which are merged or added to represent the context vector in a more compact form.

$$\mathcal{Z}_i = \mathcal{E}_i v_i \qquad (41)$$

$$C = \sum_{i=1}^{d_k} \mathcal{Z}_i \qquad (42)$$

The vectors obtained from both the attention mechanisms are further added and passed through the subsequent section of the encoder layer which is FFN.

$$\Omega_t = f(MHA(k, q, v), C) \qquad (43)$$

Also, the decoder structure follows the traditional transformer structure [48] with a unified attention mechanism. The embeddings input to the decoder network are shifted right version of the embeddings at the encoder network. These embeddings are made to pass through the multi-head attention network and the unified attention network. The dumped results are then fed through a series of FC layers and finally a linear activation function followed by a softmax layer, generating the output probabilities. In addition to this, the whole model incorporates a pointer generator network [43] and a coverage mechanism [43]. The context vector is obtained from the encoder output and the output state of the hidden layer inside the decoder and predicts new words for the dictionary. The pointer network's attention provides information about the most significant words at any given time, which is helpful for prediction. The current moment's attention distribution and the sum of weights for the encoder's hidden layer are represented as:

$$b^t = softmax(\Omega_t) \qquad (44)$$

$$n_t = \sum_i b^t h_i \qquad (45)$$

Further, vocabulary probability distribution and the probability of distribution of words is mathematically expressed as:

$$\mathcal{P}_{vocab} = softmax(t'(t[\Omega_t, n_t] + u) + u') \qquad (46)$$

Where, $t, t', u, u'$ are the learnable parameters. Also, the probability distribution of words is:

$$\mathcal{P}(\mathcal{W}) = \mathcal{P}_{vocab}(\mathcal{W}) \qquad (47)$$

Hence, the model through a pointer network must either directly copy words from the source or create new terms to address the out-of-vocabulary issues. Therefore, the pointer-generator network is employed with probability to copy a word from the source text or to generate the word:

$$\mathcal{P}_{gen} = sigmoid(w_h^T h_t^* + w_n^T n_t + w_x^T x_t + r_{ptr}) \qquad (48)$$

Where, $w_n^T, w_h^T, w_x^T$, and $r_{ptr}$ are learnable parameters and $\mathcal{P}_{gen} \epsilon [0,1]$.

Further, $\mathcal{P}_{gen}$ acts as a switch that either generates a word from the vocabulary using sampling from $\mathcal{P}_{vocab}$, or copies a word from the input sequence by sampling from the attention distribution. Also, there exists an extended vocabulary for each document which union of the vocabulary, and all words appearing in the source document.

$$\mathcal{P}(\mathcal{W}) = \mathcal{P}_{gen} \mathcal{P}_{vocab}(\mathcal{W}) + (1 - \mathcal{P}_{gen}) \sum_{j:w_j} a_{ij} \qquad (49)$$

Note, if $w$ is an OOV word, then $\mathcal{P}_{vocab}(\mathcal{W}) = 0$. Also, if $w$ does not appear in the source document,

then $\sum_{j:w_j} a_{ij} = 0$.

For time-step $i$, negative log-likelihood loss is evaluated for the target word $w_i$ and is given by:

$$\mathcal{L}_i = -log\mathcal{P}(\mathcal{W}_i) \qquad (50)$$

$$\mathcal{L} = \frac{1}{T}\sum_{i=0}^{T}\mathcal{L}_i \qquad (51)$$

where $T$ is the target sequence length. To overcome the issues related to the repetition of words, a coverage mechanism is employed to direct the attention version to non-repeating words. The attention distributions generated using all previous prediction steps are added to create the coverage vector $\chi_i^v$.

$$\chi_t = \sum_{t'=0}^{t-1} \Omega_{t'} \qquad (52)$$

Therefore, equation (43) is modified by considering the effect of the coverage mechanism. Hence, this makes it easier for the attention mechanism to stop attending to the same places repeatedly and stop producing repetitive text as a result.

$$\Omega_{t'} = f(MHA(\hbar, q, \mathcal{v}), C, \chi_t) \qquad (53)$$

The coverage loss is also defined to penalize repeatedly attending to the same locations.

$$\mathcal{L}_c = \lambda \sum_i \min(\chi_t, \Omega_{t'}) \qquad (54)$$

where $\lambda$ is the balancing parameter. Hence, the total loss for the proposed $UnMHA - ST$ is the combination of equations (53) and (54) respectively,

$$\mathcal{L}_{total} = \mathcal{L} + \mathcal{L}_c \qquad (55)$$

## 4. RESULTS AND DISCUSSIONS

To validate the effectiveness of the proposed MAA-FIC experiments are conducted on the Flickr8K dataset. This dataset contains around 8K images paired with five descriptions for each image. To generate a style-based description of an image FlickrStyle10K dataset is utilized. This dataset contains textual annotations for romantic and humorous styles respectively. For this dataset, only 7K annotations are publicly available. For the task of summarization, the text data obtained from the two captioning methods (discussed in Section III) is collected and utilized for the generation of summarized description of an image reflecting factual, romantic, and humorous elements. The data used for summarization contains around 7000 paragraphs with three sentences (one each for factual, romantic, and humorous caption).

### 4.1. Implementation Details

For the factual caption generation model, Adam [53] optimizer is utilized for the minimization of cross-entropy loss with a learning rate of $2e - 5$. Further, the batch size is set as 64 and the proposed model is trained for 70 epochs. Also, to extract. the text features, fine-tuned GloVe embeddings are utilized with embedding size as 300. To evaluate the performance of the proposed factual image captioning model, BLEU@N [55] and METEOR [56] scores are evaluated.

To generate romantic and humorous captions, ADAM [57] optimizer is utilized at the encoder end with a learning rate of $2e - 5$. For language decoder embedding size is set as 300 and the dimensions of the hidden layer of the proposed SF-Bi-ALSTM module is set as 512. The whole model is trained for about 60 epochs with a batch size of 96. To evaluate the performance of generated stylized captions, style transfer accuracy and perplexity of the proposed model are evaluated. Also, the relevancy of the captions generated is evaluated in terms of BLEU@N [55] and METEOR [56].

For the summarization module, a single-layer $UnMHA - ST$ transformer with a pointer generator network and coverage mechanism is incorporated. The model is trained for 100 epochs with a dropout rate of 0.1 and a batch size of 4. Further, the pointer-generator network leverages 300-dimensional GloVe embeddings with a vocabulary size of 400K. To evaluate the performance of the proposed model ROUGE-1, ROUGE-2, and ROUGE-L scores are evaluated.

### 4.2 Experimental Results

This section presents the results for (i) Modified Adaptive Attention-based Factual Image Captioning (MAA-FIC) Model, (ii) Style-based caption generation model, and (iii) Caption Summarization model.

*4.2.1 MAA- FIC Model*

Table I reports the comparison of the proposed MAA-FIC model with other state-of-the-art on the Flickr8K dataset. The comparison of the proposed model is made in terms of *BLEU@N* and *METEOR* scores. MAA-FIC utilized fusion of Inception-V3 and Faster R-CNN at the encoder and provided provides state-of-the-art results when compared with ImageNet-CNN [21], VGGNet-CNN [20] based feature extraction method. Also, Jia et al. [58] utilized semantic information from the image as an extra input to each unit of the LSTM block. Very few works [1] [6] [26] focused on attention mechanism-based architectures and the proposed MAA-FIC utilized modified adaptive attention (MAA). MAA combines channel and spatial attention mechanisms to focus more on most of the important image information as well as the position of the image regions. Furthermore, the MAA provided a significant increase in the BLEU@N and METEOR scores when compared with the [59] [24]. Fig. 6 (a) and (b) depict the accuracy and loss curves for the proposed MAA-FIC. The curves make it evident that with the increase in the number of epochs, the accuracy of the proposed model increases while the losses decrease. Fig.7 represents the qualitative results obtained for the proposed MAA-FIC model. The factual captions obtained for Flickr8K produces syntactically and semantically correct descriptions by focusing more on the salient object regions. Also, by leveraging the modified adaptive attention the proposed model provides a better correlation between the objects. However, there are a few instances where the proposed MAA-FIC model could not detect the correct object. For example, in Fig. 7(b) the MAA-FIC model predicted "*mobile*" instead of "*camera*".

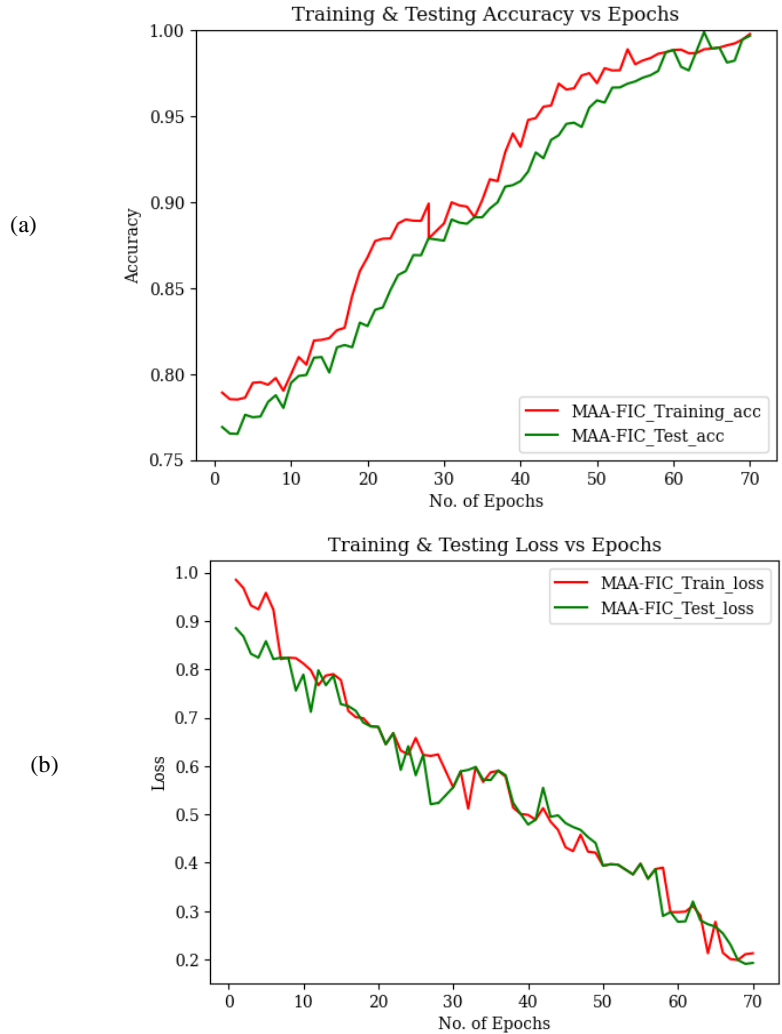

(a)

(b)

**Fig. 6:** (a) Accuracy and (b) Loss curves for the proposed MAA-FIC

**TABLE I**
**RESULTS ON FLICKR8K DATASET**

| Model | B-1 | B-2 | B-3 | B-4 | M |
|---|---|---|---|---|---|
| LSTM [21] | 66.0 | 42.0 | 27.0 | 18.0 | - |
| g-LSTM [58] | 64.7 | 45.9 | 31.8 | 21.6 | 20.60 |
| Log Bilinear [59] | 65.6 | 42.4 | 27.7 | 17.7 | 17.31 |
| SCA-CNN [6] | 68.2 | 49.6 | 35.9 | 25.8 | - |
| Hard Attention [1] | 67.0 | 45.7 | 31.4 | 21.3 | 20.30 |
| m-RNN [20] | 48.2 | 35.7 | 26.9 | 20.8 | - |
| AA + Bi-LSTM [26] | 70.2 | 49.1 | 35.9 | 26.6 | - |
| Deep-Bi-LSTM [24] | 65.5 | 46.8 | 32.0 | 21.5 | - |
| SDCD [60] | 67.2 | 45.1 | 30.5 | 21.5 | - |
| JRAN [61] | 67.7 | 47.4 | 33.2 | 22.7 | 20.9 |
| **MAA-FIC** | **74.8** | **52.3** | **38.7** | **26.8** | **22.6** |

### 4.2.3 SF-Bi-ALSTM Based Stylized Image Captioning

Table II reports the results using both the romantic references and the humorous references for the proposed SF-Bi-ALSTM-based stylized image captioning module with other state-of-the-art. The comparison is carried out with respect to $BLEU-1$, $BLEU-3$, and $METEOR$ ($M$) scores with style accuracy ($cls$) and perplexity ($ppl$). The results reported in Table II make it evident that (1) given a specified style, the proposed model is tailored to that style outperforming the baseline approaches across different automatic evaluation metrics; (2) the relative performance variation shows that the proposed model can successfully simulate the style factors in caption generation. Further, StyleNet [11] and SF-LSTM [9] incorporate factored LSTM and style-factual LSTM decoders for the generation of stylized captions. These methods proposed end-to-end learning. framework for the generation of factual and style-based descriptions. On the other hand, the proposed model defines a modified style factored Bi-LSTM with captions. Further, the works [10] [12] implemented the task of multi-style-based caption learning using unpaired data but these methods when compared with the proposed stylized captioning model generate more attractive and coherent style-based descriptions.

TABLE II
RESULTS ON FLICKRSTYLE10K DATASET

| Style | Model | B-1 | B-3 | M | cls | ppl |
|---|---|---|---|---|---|---|
| Romantic | StyleNet [9] | 13.3 | 1.5 | 4.5 | 57.1 | 6.9 |
| | MSCap [12] | 17.0 | 2.0 | 5.4 | 91.3 | - |
| | SF-LSTM [11] | 27.8 | 8.2 | 11.2 | - | - |
| | SAN [10] | 28.3 | 8.7 | 11.5 | 90.9 | 9.1 |
| | Detach and attach [62] | 24.3 | - | - | 82.4 | - |
| | Wu et al. [63] | 25.4 | 5.7 | 9.2 | - | - |
| | Proposed | **29.8** | **9.2** | **11.9** | **92.0** | **9.3** |
| Humorous | StyleNet [9] | 13.4 | 0.9 | 4.3 | 42.5 | 7.3 |
| | MSCap [12] | 16.3 | 1.9 | 5.3 | 88.7 | - |
| | SF-LSTM [11] | 27.4 | 8.5 | 11.0 | - | - |
| | SAN [10] | 27.6 | 8.1 | 11.2 | 87.8 | 8.4 |
| | Detach and attach [62] | 23.0 | - | - | 89.2 | - |
| | Wu et al. [63] | 27.2 | 5.9 | 9.0 | - | - |
| | Proposed | **29.2** | **9.0** | **11.8** | **89.8** | **8.7** |

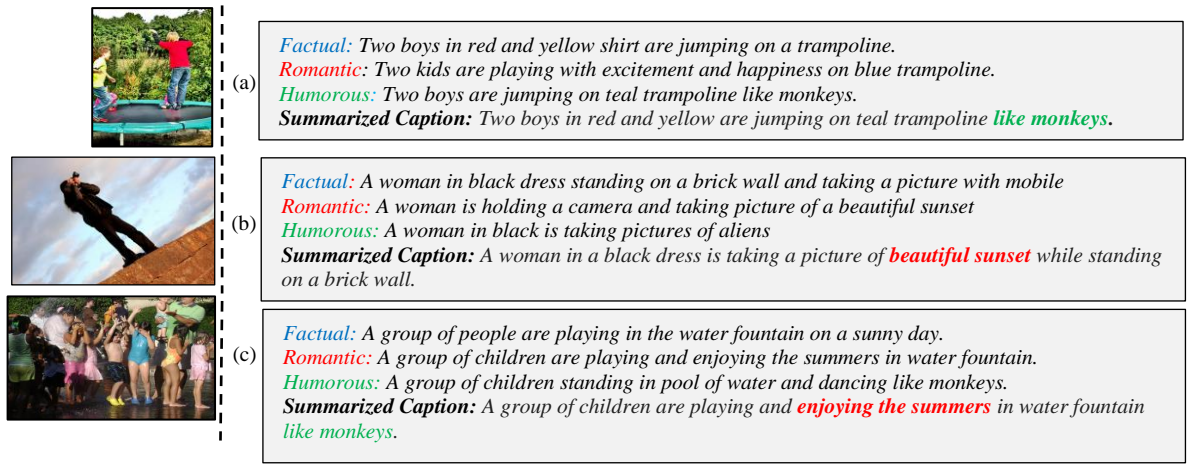

**Fig. 7:** Qualitative results obtained for the proposed $UnMA-CapSumT$ (a) UnMA-CapSumT Factual + Humorous (b) UnMA-CapSumT Factual + Romantic (c) UnMA-CapSumT Factual + Romantic + Humorous

The qualitative results presented in Fig.7 for the generation of romantic and humorous proves that the generated descriptions describe the content of the image well. The generated captions express the content in a romantic *(A little girl enjoying the joys of childhood with her brother in background)* or humorous *(A group of children standing in pool of water and dancing like monkeys).* More intriguingly, the descriptions generated are not only romantic or humorous but they also suit the visual content of the image coherently, making the caption visually appealing and relevant. Also, qualitative results were obtained to verify that the sentences so generated provide a strong correlation with the human evaluation.

### 4.2.4 UnMHA-ST Text Summarization Transformer

The summarized captions generated from the proposed UnMHA-ST is presented in Fig. 7. From Fig. 7 it is evident that there exist some cases as shown in Fig. 7(a), according to the context of the image the summarized captions depict only the factual and humorous elements and for a few images as shown in Fig. 7(b) romantic style has dominated with the factual element. Further, there are cases (Fig. 7(c)) in which the proposed summarization framework can successfully include romantic and humorous styles with factual elements.

**TABLE III**
**ABLATION RESULTS OBTAINED FOR SUMMARIZED CAPTIONS**

| Model | Transformer | Embedding | R-1 | R-2 | R-L |
|---|---|---|---|---|---|
| Model-1 | Baseline [48] | Doc2Vec | 20.18 | 4.33 | 13.71 |
| Model-2 | Baseline [48] | GloVe | 21.3 | 4.43 | 14.19 |
| **Model-3** | **Baseline [48]** | **fastText** | **23.46** | **4.65** | **14.77** |
| Model-4 | $UnMHA-ST$ | fTA-WE | 26.72 | 5.71 | 15.24 |
| Model-5 | $UnMHA-ST$ + Pointer Generator Network | fTA-WE | 28.32 | 6.43 | 17.72 |
| **Model-6** | $UnMHA-ST$ + **Pointer Generator Network + Coverage Mechanism** | **fTA-WE** | **30.11** | **6.73** | **18.22** |

Table III reports the ROUGE-1, ROUGE-2, and ROUGE-L scores for the baseline transformer model [48] with Doc2Vec, GloVe, and fastText word embeddings. From the results reported it is evident that the fastText word embeddings are superior as they can derive word vectors for unknown words or out of vocabulary words. Further, the baseline transformer model is replaced with the proposed $UnMHA-ST$ and fTA- WE word embedding which results in significant improvement of R-1, R-2, and R-L scores. To overcome the OOV issues the proposed $UnMHA-ST$ transformer is utilized with pointer-generator and to further enhance the performance of the proposed framework and also to avoid the repetition issues $UnMHA-ST$ is equipped with coverage mechanism. This provides significant improvements by generating fluent summarized captions. Also, box plot-based model performance

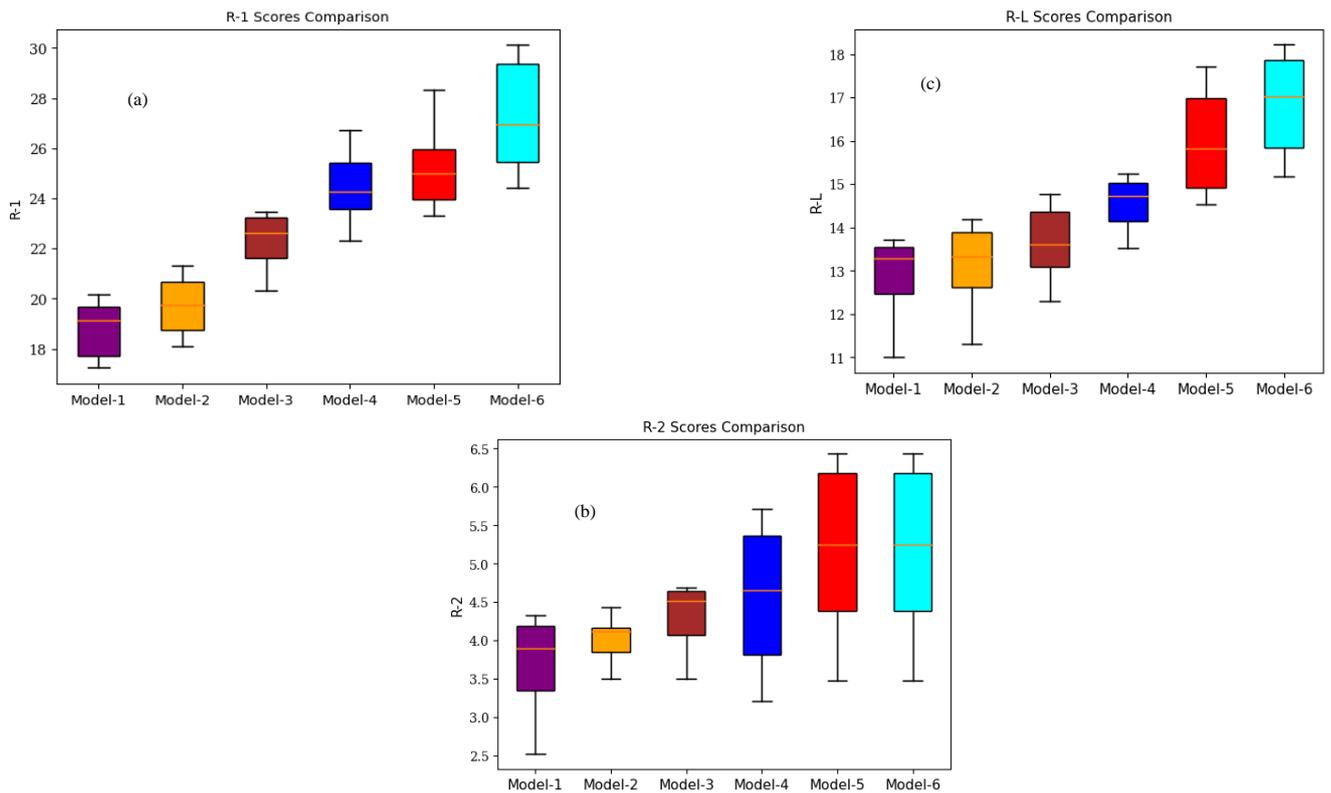

**Fig. 8:** Comparison for Quantitate Results Obtained for the proposed Framework (a) R-1, (b) R-2, and (c) R-L

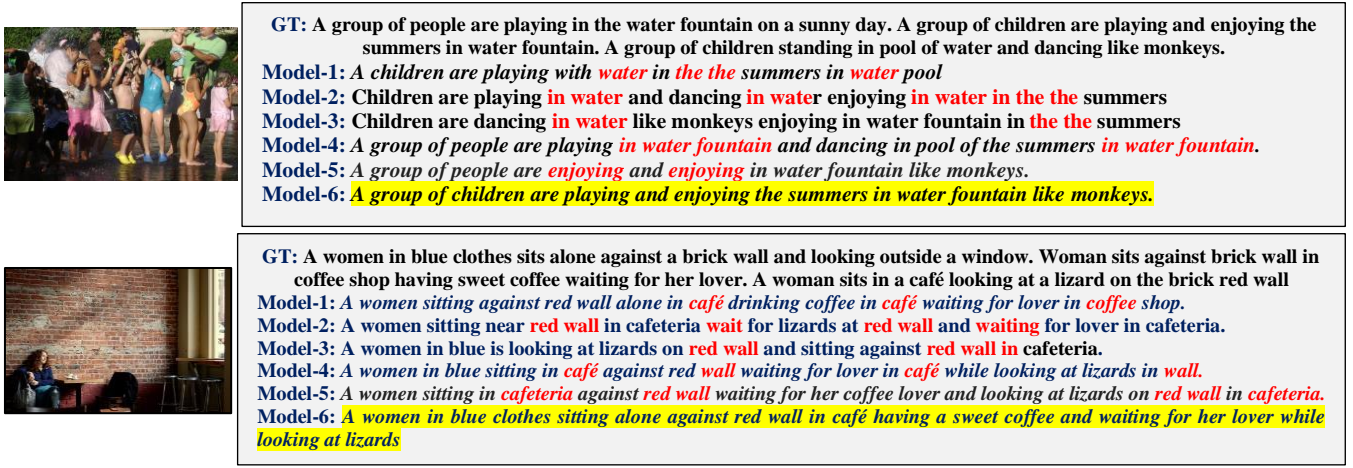

**Fig.9:** Observed Qualitative Results for the proposed *UnMA*-CapSumT *(Text in Red represents the repeated words that appeared in the summarized captions)*: Model -6 (proposed model) provides a summarized caption for an image without repetition of words and out-of-vocabulary issues

comparison in terms of R-1, R-2, and R-L is shown in Fig. 8.

Further, Fig. 9 presents the qualitative ablated summarized captions for Model 1 to Model-6. Model-1, the baseline transformer model with Doc2Vec word embedding generated a poor-quality summarized caption with redundant and repeated information. Further, fastText embedding with the baseline transformer model (Model-3) provided a slight improvement in the generated summarized captions. With the use of the proposed $UnMHA-ST$ and fTA-WE in Model-4 provided an improvement in results but still there exists repetition of words like "in water fountain" in first case and "café" and "wall" in second case. In Model-5, the pointer-generator network adaptively points to the contextual words with appropriate styles hence resulting in generation of romantic and humorous elements in the summarized captions. For the second case, Model-5 generates a word "cafeteria" that is synonym to the word "café" or "coffee shop". Further, Model-6 generated a summarized caption that provided a strong correlation with the human evaluation by generating syntactically and semantically correct descriptions by highlighting both factual and stylized elements. Also, the summarized caption generated is free from OOV and repetition problems with the use of a pointer-generator network and coverage mechanism.

## 5. CONCLUSION

This paper presents a novel caption summarization framework, $UnMA-CapSumT$ to generate summarized captions highlighting the factual, romantic, and humorous elements in a single caption. The proposed framework is divided into two stages: $(i)$ generation of factual, romantic, and humorous descriptions of images using the MAA-FIC model and SF-Bi-ALSTM-based stylized image captioning model, $(ii)$ using the descriptions generated in $(i)$ to produce a summarized single line caption for the images by incorporating multi-head attention and unified attention driven $UnMHA-ST$ transformer. The proposed $UnMHA-ST$ transformer utilizes a pointer-generator network and coverage mechanism to avoid the issues related to OOV and repetition problems. The summarized caption generated provides an improvement in learning knowledge of factual content and its associated linguistic styles. Also, the observed qualitative results make it evident that the proposed framework provided a strong correlation with human evaluation by generating semantically and syntactically correct descriptions. Further, the improvements in the performance and interpretability of the proposed framework are enhanced with the utilization of an efficient word embedding fTA-WE. To validate the effectiveness of the proposed framework extensive experiments are conducted on Flickr8K and FlickrStyle10K. Experimental results prove that the proposed MAA-FIC, SF-Bi-ALSTM-based image captioning model, and $UnMHA-ST$ summarization transformer model provided state-of-the-art results and generates visually and grammatically correct factual, romantic, humorous, and summarized captions for a given image.